# Clustering-based Anomaly Detection in Multivariate Time Series Data

Jinbo Li[a], Hesam Izakian[a], Witold Pedrycz[a,b], and Iqbal Jamal[c]

[a] Department of Electrical & Computer Engineering
University of Alberta
Edmonton T6R 2V4 AB Canada
[b] Warsaw School of Information Technology, Newelska 6
Warsaw, Poland
[c] AQL Management Consulting Inc., Edmonton, AB, Canada

**Abstract**
Multivariate time series data come as a collection of time series describing different aspects of a certain temporal phenomenon. Anomaly detection in this type of data constitutes a challenging problem yet with numerous applications in science and engineering because anomaly scores come from the simultaneous consideration of the temporal and variable relationships. In this paper, we propose a clustering-based approach to detect anomalies concerning the amplitude and the shape of multivariate time series. First, we use a sliding window to generate a set of multivariate subsequences and thereafter apply an extended fuzzy clustering to reveal a structure present within the generated multivariate subsequences. Finally, a reconstruction criterion is employed to reconstruct the multivariate subsequences with the optimal cluster centers and the partition matrix. We construct a confidence index to quantify a level of anomaly detected in the series and apply Particle Swarm Optimization as an optimization vehicle for the problem of anomaly detection. Experimental studies completed on several synthetic and six real-world datasets suggest that the proposed methods can detect the anomalies in multivariate time series. With the help of available clusters revealed by the extended fuzzy clustering, the proposed framework can detect anomalies in the multivariate time series and is suitable for identifying anomalous amplitude and shape patterns in various application domains such as health care, weather data analysis, finance, and disease outbreak detection.

# 1. Introduction

In many real-world applications including load demand forecasting [1], geography [2], human activity recognition [3], stock return [4] and others [5], collected data arise in the form of time series. Anomaly detection in this type of data refers to the discovering of any abnormal behavior within the data encountered in a specific time interval. Anomaly detection has been widely used in numerous application areas. For instance, cardiologists are interested in identifying anomalous parts of ECG signals to diagnose heart disorders [6]. Economists are interested in anomalous parts of share prices to analyze and build economic models [7]. Meteorologists are interested in anomalous parts of weather data to predict future consequences [8]. Although numerous time series anomaly detection techniques have been reported in the literature, see e.g., [9, 10], most of these techniques are concerned with univariate time series.

In this study, we propose a clustering-based approach for anomaly detection in multivariate time series. Detecting anomalous parts of multivariate time series constitutes a challenging problem. The reason is that the temporal relationship between time points has to be taken carefully into account along with the relationships present among individual time series [11]. Consequently, the technique discussed for this purpose has to consider all variables at the same time to determine an anomaly score [12].

Here, we consider two common types of anomalies [13], namely anomaly in amplitude and shape [13] and propose a clustering-based multivariate time series anomaly detection technique. In the first step of the method, a fixed-length sliding window is applied to generate a set of multivariate subsequences. Then, using an extended version of the Fuzzy C-Means clustering [14], the available (normal) structure within the generated multivariate subsequences is discovered. Finally, using a reconstruction technique, an anomaly score quantifying the level of departure (anomaly) from the normal structure of data is assigned to each multivariate subsequence.

When dealing with amplitude anomalies, the original feature space along with the Euclidean distance is employed in the clustering process. As to anomalies in shape, the autocorrelation representation of time series is used to capture shape information and remove time shifts in data. Then, the Euclidean distance function is employed to quantify the similarity/dissimilarity of time series data in the new feature space.

Overall, the main contributions of this paper can be outlined as follows:

1) Developing an extended version of the Fuzzy C-Means clustering to reveal the available structure within multivariate time series data.
2) Developing a general framework for anomaly detection (both in shape and the amplitude) in multivariate time series.

The originality of this study exhibits three essential facets: (i) We offer a unified framework for detecting anomalous segments of data with respect to the amplitude and/or shape information in multivariate time series data. (ii) Detecting anomalous parts of multivariate time series data with the use of available clusters within the data is a novel idea proposed here. (iii) The underlying originality of the extended Fuzzy C-Means clustering technique relies on assigning different weights to the different components of the time series to control the impact of each variable in the clustering process of multivariate time series.

This study is organized as follows. A brief overview of time series anomaly detection techniques is reported in Section II. The detailed algorithmic aspects, including clustering, correlation analysis, and a reconstruction process developed for multivariate time series, are introduced in Section III. Experimental studies completed for both synthetic and real-world multivariate time series are reported in Section IV. Finally, we conclude the paper in Section V.

## 2. Literature review

Various anomaly detection techniques for univariate time series data have been proposed in the literature (refer e.g., to [12, 15]). Comparing to univariate time series, anomaly detection in multivariate time series has been more challenging since more than a single variable have to be considered simultaneously when detecting anomalous segments of data. Methods of anomaly detection in time series data can be divided into a set of categories, namely similarity-based methods [16], clustering-based methods [17], classification-based methods [18, 19], modeling-based methods [15], frequency-based methods [20], and probability-based methods [21, 22]. A thorough survey of time series anomaly detection techniques is reported in [9, 12]. In what follows, we briefly recall the main features of these approaches.

### 2.1. Similarity-based methods

A simple technique to determine anomalies in time series is to use a similarity measure along with a brute force algorithm [23]. Generally, subsequences exhibiting high differences from the other subsequences are regarded as anomalies based on the similarity measurement. In similarity-based techniques, selecting a suitable similarity measure might have a strong impact on the performance of the method and directly depends on the application purpose [24]. For instance, when the time series are collected at different sampling rates, Dissim distance [25] can be considered since it uses a finite set to define a time series. Dynamic time warping distance (DTW) [26, 27] is another effective technique that can be considered when the lengths of time series are unequal and there are temporal shifts in data. Some other similarity measures such Longest common subsequence (LCSS), Edit Distance on Real sequence (EDR) and alike are widely reported in the literature [28]. Another approach [29] to detect anomalies contains two main steps: firstly, employ sparse coding to extract features, and then use latent semantic analysis (LSA) to learn relationship. Finally, the squared reconstruction errors are considered as anomaly scores. Note that the reference time series used in this method should be without any abnormal subsequences. In [30], a linear method for anomaly detection is proposed; it demonstrates the advantages of low time complexity and a lower number of parameters. However, only the top discords $k$ can be detected and there is not detailed guidance to determine the preferred value of $k$.

## 2.2. Clustering-based methods

Applying clustering methods on multivariate time series is another option. In general, the clustering process of clustering based techniques can be divided into two main phases [31]. In order to generate subsequences of multivariate time series, the first stage is to implement a clustering method such as *K*-Means [32] or Fuzzy C-Means (FCM) [33] over the time series. Other clustering techniques can also be considered as well [34, 35]. At the second stage, they determine an anomaly score on a basis of some measurement of the fitness of subsequences [23] to the different clusters. The anomaly score is assigned in a simple way by using the distance between the subsequences and the cluster centers. In [36], the weighted Euclidean distance between the observations is computed and then an improved ant colony method [37] is exploited to complete clustering. In [38], a novel clustering based compression method is proposed to reduce the time complexity of the approach. *K*-Means algorithm is utilized to convert a correlation matrix of multivariate time series to a matrix of multiple clusters. According to the multivariate normal distributions, the anomaly scores were estimated. In [39], a Bounded Coordinate System (BCS) approach [40] is considered to evaluate the similarity between two multivariate time series. A modified version of *K*-Means is employed to cluster the multivariate time series dataset. The top $k$ outlier observations were detected using a two-pruning rule-based nested loop algorithm. The benefit of the clustering-based models is that prior knowledge of the data is not necessary. However, the clustering-based methods suffer from the fact that the revealed clustering centers have an essential impact on the performance of the detection models and high time and space complexity. If the structure within the multivariate time series cannot be revealed, the accuracy of anomaly detection will deteriorate.

## 2.3. Classification-based methods

In classification-based methods, instances or subsequences of multivariate time series are classified into two classes: normal and abnormal. The classifier is trained through a training set composed of normal instances and then it can be used to assign an anomaly score to each instance of testing set [41]. In [22], the authors proposed to determine whether a data point is abnormal by exploiting the linear regression model firstly and then learn a Bayesian maximum likelihood classifier on the basis of anomalies identified. For the testing dataset, the classifier labels each test data point. The authors used the Meta-Features (i.e., Kurtosis, variation, oscillation, regularity, square, and trend) to describe the dynamics of subsequences and the support vector machine to detect the anomalies [42]. The advantage of the classification-based model is that the detection accuracy is relatively higher than that of unsupervised detectors in most cases. However, for classification-based methods, collecting training data that are used to build the classifiers is known to be a time-consuming process.

## 2.4. Transformation-based methods

An option in dealing with multivariate time series anomaly detection is to reduce the multivariate time series to univariate time series. Then a univariate anomaly detection technique can be employed to detect anomalies. Two common transformation methods are time series projection and independent component analysis [43]. The objective is to reduce the dimensionality of multivariate time series and speed up the computation. However, it should be noted that the loss of information in the process of transformation might reduce the accuracy of the multivariate time series anomaly detection [11]. Parthasarathy et al. [44] proposed a novelty dissimilarity measurement for comparing multivariate time series data based on principal component analysis and provided a point anomaly detection algorithm in multivariate time series. The dissimilarity measurement contains distance (Euclidean or Mahalanobis distances), rotation and variance components. The value of dissimilarity was calculated by combining these three components. In order to add potential domain knowledge and improve the flexibility of the method, some coefficients were added into the method to allow potential users to assign different weights to the different components of the time series. How much useful information the transformation methods can retain is essential for the performance of the transformation-based detectors.

## 2.5. Modeling-based methods

Multivariate time series modeling is another anomaly detection method reported in the literature. In [15], the authors combined Hidden Markov Model (HMM) and different transformation approaches (i.e., PCA, FCM, Sugeno Integral, and Choquet Integral) to perform multivariate time series anomaly detection. The first stage is similar to the idea of the transformation-based model and to retain useful information. In the next stage, HMM is used to model the temporal relationship. The drawback of the method is that the supposition (e.g., two finite sets of states such as visible and hidden states always exist) is not suitable for data with more complex structure. Qiu al. et. [45] proposed to learn graphical models of Granger causality and then exploit Kullback-Leibler (KL) divergence to compute the anomaly score for each variable. A threshold that is from reference data determines whether the observation is abnormal. The paper proposed a multi-scale convolutional recurrent encoder-decoder to detect the anomalies, which used the resolution signature matrices and convolutional long-short term memory network to capture the characteristics and temporal patterns, respectively [46]. Cheng al. et [11] proposed the use of a weighted graph representation to model multivariate time series. RBF function was employed to obtain the similarity between a pair of multivariate time series. In the graph, nodes were subsequences or observations and edges were considered as the similarity between nodes. Random walk algorithm [47] was used to produce the connectivity values. Then the connectivity measure of nodes was considered to detect anomalies. However, for most existing modeling-based methods, their performance will strongly depend on the observed data [48]. Moreover, the reference data are required to build models. Overall, based on various applications, definitions and background from the literature, there are different ways to formulate the problem of time series anomaly detection. The related techniques can be grouped into supervised and unsupervised categories according to whether the reference or training time series is available. Collecting and annotating training data is highly time-consuming and not very practical for most real-world applications because domain-specific knowledge is required. For unsupervised approaches, a suitable similarity measure is essential for both the similarity-based methods and the clustering-based methods. The performance of clustering-based methods heavily depends on the available structure of time series captured by the clustering methods.

## 3. The proposed approach

In this section, we introduce a clustering-based approach for anomaly detection in multivariate time series data. Figure. 1 and 2 display an overall flow of processing carried out by running the methods for detecting anomalies in amplitude and shape, respectively. First, a fixed-length sliding window (in both Figure 1 and 2) is employed to divide the long multivariate time series into a set of multivariate shorter subsequences. In the next step of this proposed framework, an extended version of the FCM clustering is used to reveal the available structure within the data. Thereafter, a reconstruction criterion along with a particle swarm optimization is considered. Finally, the anomaly score of each subsequence is assigned based on the revealed clusters and the reconstruction criterion. Compared with the proposed amplitude anomaly detection method, there is an extra component for shape anomaly detection (see Figure. 2). In fact, before performing the anomaly detection with respect to shape information, an autocorrelation representation of time series has been considered to capture shape information and remove time shifts within data. In what follows, we describe each component of the proposed method in a more formal way.

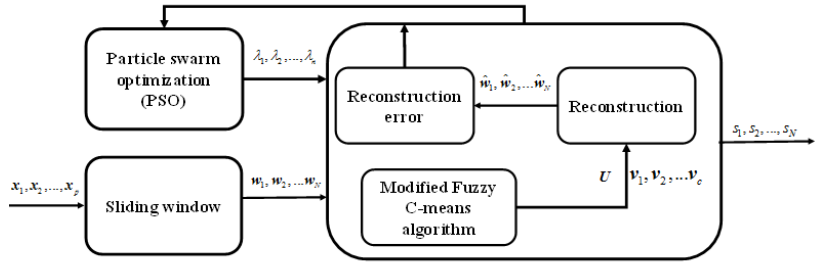

Figure 1 Overall scheme of anomaly detection in amplitude

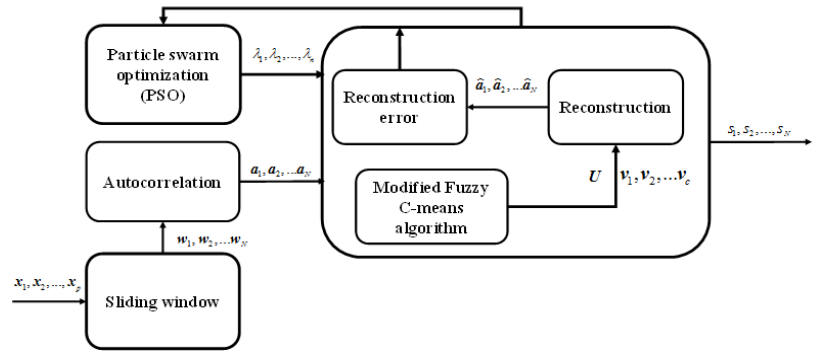

Figure 2 Overall scheme of anomaly detection in shape

### 3.1. Sliding window

Let us assume that $x_1, x_2, \ldots, x_p$ are multivariate time series of length $p$. The $k^{\text{th}}$

point in this time series is expressed using $n$ variables, viz. $\boldsymbol{x}_k = [x_{k1}, x_{k2}, \ldots, x_{kn}]$, where $n$ is the number of variables present in the multivariate time series. Using a fixed-length sliding window, one may generate a set of $N$ subsequences of length $q$. Figure. 3 shows an example of generating multivariate subsequences using the sliding window technique. The sliding window moves across the time series at each movement, and the data inside the sliding window is considered as a subsequence. The first and second subsequences are visualized in the Figure 3. Assuming $r$ to be the length of each movement of the sliding window, the number of generated subsequences can be determined as follows

$$N = \frac{p - q}{r} + 1 \tag{1}$$

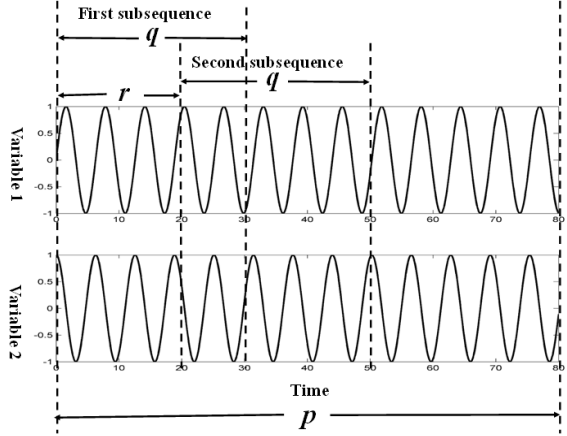

Figure 3 The use of sliding window to generate multivariate subsequence.

### 3.2. An extended Fuzzy C-Means for clustering multivariate time series

The sliding window component generates a set of multivariate subsequences. After obtaining the subsequences, the FCM is used to cluster them to reveal the available structure within the data. Since that each multivariate subsequence consists of two or more univariate subsequences, and these univariate subsequences may have different characteristics and structures, clustering this type of data using a standard FCM technique may lead to bias towards one or more variables in the data.

To deal with this issue, we introduce a novel extended version of fuzzy clustering for multivariate time series (subsequences). Here the extended version of the Euclidean distance function has been considered to control the impact of each variable in evaluating the similarity between multivariate time series. In the proposed extended distance function, the squared Euclidean distance between a multivariate subsequence $\boldsymbol{w}_j$ and a cluster center $\boldsymbol{v}_j$ can be calculated as follows:

$$d^2(\boldsymbol{w}_j, \boldsymbol{v}_i) = \lambda_1 \|w_{j1} - v_{i1}\|^2 \\ + \cdots + \lambda_n \|w_{jn} - v_{in}\|^2, \lambda_i \geq 0, \sum_{i=1}^{n} \lambda_i = 1 \tag{2}$$

Using the distance function in (2), one may control the impact of each variable in the clustering process of multivariate time series. A higher value of $\lambda_i$, leads to the increase of the impact of the $i^{\text{th}}$ variable and the decrease of the impact of the other variables in the clustering process. By inserting the extended distance function in FCM objective function, we obtain at the following expression.

$$Q = \sum_{i=1}^{c} \sum_{j=1}^{N} u_{ij}^m d^2(\mathbf{w}_j, \mathbf{v}_i) \quad (3)$$

where $c$ is the number of clusters, $m(m > 1)$ is the fuzzification coefficient, and $N$ is the number of multivariate subsequences. $U$ and $\mathbf{v}_i$ are the partition matrix and prototype $i$, respectively. The optimization process of the proposed extended objective function is realized in an iterative fashion. Here we compute the partition matrix and cluster centers using the following expressions.

$$\mathbf{v}_i = \frac{\sum_{j=1}^{N} u_{ij}^m \mathbf{w}_j}{\sum_{j=1}^{N} u_{ij}^m} \quad (4)$$

$$u_{ij} = \frac{1}{\sum_{l=1}^{c} \left(\frac{\|\mathbf{v}_i - \mathbf{w}_j\|}{\|\mathbf{v}_l - \mathbf{w}_j\|}\right)^{2/(m-1)}} \quad (5)$$

Using the proposed extended Fuzzy C-Means, one may control the impact of each variable in clustering multivariate time series data. However, an optimal impact of each variable in the clustering process is required. In other words, an optimal value for each coefficient $\lambda_i, i = 1,2,\ldots,n$ has to be estimated. To carry out the purpose, we present a reconstruction criterion along with a Particle Swarm Optimization (PSO) algorithm as a tool of searching for the optimal values. The PSO algorithm can find (near) optimal values of the weights because of its superiority in solving the complex global optimization problem by using flock of particles, which constitute a swarm and are characterized by positions and velocities. Based on the fitness function, each particle can move in the search space and store its best location of all those visited by itself in the personal memory. The global best position is determined for the entire swarm. When each particle moves to a new location, it refers both its personal best location and the global best location. The iteration of evolution intermates from converged conditions or the maximum number of iterations.

In this paper, since that there are $n$ weights $\lambda_i, i = 1,2,\ldots,n$, in the extended distance function expressed in (2), considering all possible combinations of these values (to find an optimal impact of each variable in clustering process) is time consuming and for higher values of $n$ is not even feasible. Therefore, we use PSO as a vehicle to seek for an optimal combination of the values of weights $\lambda_i$ that satisfy the constraint defined in (2). It starts with producing a number of particles and their velocity vectors randomly. For each particle, we evaluate its quality using a reconstruction criterion, which serves as a fitness function of the PSO technique. The particles and their velocity vectors are updated as follows [49].

$$v_{ki}^{t+1} = w \times v_{ki}^t + c_1 r_{1i}(pbest_{ki}^t - z_{ki}^t) + c_2 r_{2i}(gbest_i^t - z_{ki}^t) \quad (6)$$

$$z_{ki}^{t+1} = z_{ki}^t + v_{ki}^{t+1}, \quad k = 1,2,\ldots,M; i = 1,2,\ldots,n \quad (7)$$

where, $v_{ki} \in [v_{min}, v_{max}]$, $M$ is the number of particles and $n$ is the number of variables in multivariate time series. The maximum and minimum velocities are used to limit the velocity of each particle for avoiding the possible chaotic behavior of the swarm. $pbest$ and $gbest$ are best location achieved by particle and the best location found by the whole swarm, respectively. The inertia weight, $w$, controls the impact of the previous velocity on the current one and adjusts the balance between global and locale searching ability. $c_1$ and $c_2$ are the cognitive and social factors, and $r_{1i}$ and $r_{2i}$ are random values in the interval, which strikes the balance between the exploration and exploitation search. While applying the PSO algorithm to different optimization scenarios, different parameters mentioned before play an essential role in determining PSO performance. Slow convergence and poor performance may be caused by the improper choice of these parameters. In this study, we utilize the PSO as the optimizer to find the optimal combination of weights and limit ourselves to the use of static parameters which have been used in published papers and found that their performance was sufficient. To reduce the impact of the user on the performance, the parameter settings-free versions of PSO, which are achieved by following different strategies of considering parameters and solving numeric optimization problems, could be integrated into our methods in the future. For instance, the authors proposed the self-adaptive PSO based on the alpha-stable distribution (which can improve the global search ability) and dynamic fractional calculus (which is used to speed up the convergence) in [50]. More details about the parameter settings-free versions of PSO can be found in [51, 52].

### *3.3. Reconstruction criterion*

As discussed in the previous subsection, a reconstruction criterion serves as a fitness function considered in the PSO method [53]. Specifically, using this technique one may evaluate the quality of clusters in terms of data granulation and degranulation. The essence of the reconstruction criterion is to reconstruct the original data (subsequences) through the revealed cluster centers and their membership values. Considering cluster centers and partition matrix generated using the extended FCM, one may reconstruct the original subsequences through minimizing the following sum of distances:

$$F = \sum_{i=1}^{c}\sum_{j=1}^{N} u_{ik}^m d^2(\widehat{\boldsymbol{w}}_j, \boldsymbol{v}_i) \tag{8}$$

where, $\widehat{\boldsymbol{w}}_j$ is the reconstructed version of $\boldsymbol{w}_j$. By zeroing the gradient of $F$ with respect to $\widehat{\boldsymbol{w}}_j$, one has

$$\widehat{\boldsymbol{w}}_i = \frac{\sum_{j=1}^{c} u_{ij}^m \boldsymbol{v}_j}{\sum_{j=1}^{c} u_{ij}^m} \tag{9}$$

After reconstructing all data points using (9), we calculate the reconstruction error as the following sum of distances:

$$E = \sum_{j=1}^{N} \|\boldsymbol{w}_j - \widehat{\boldsymbol{w}}_j\|^2 \tag{10}$$

A lower value of $E$ indicates a higher quality of clusters in terms of the data granulation and degranulation. The PSO technique described in the previous sub-section, optimizes the coefficients $\lambda_i, i = 1,2,\ldots,n$ by minimizing the reconstruction error expressed in (10).

### 3.4. Reconstruction error as anomaly score

The optimal weights of the extended FCM technique can result in the optimal impact of different variables in the clustering process. The available structure of multivariate time series is presented by using the revealed cluster centers and partition matrix corresponding to optimal weights. Therefore, using the revealed optimal cluster centers and partition matrix, one may assign an anomaly score to each multivariate subsequence. For this purpose, the anomaly score for each multivariate subsequence is calculated as the squared Euclidean distance between the subsequence and its reconstructed version. Formally, an anomaly score for subsequence $\boldsymbol{w}_j$ is calculated in the following form.

$$s_j = \left\| \boldsymbol{w}_j - \widehat{\boldsymbol{w}}_j \right\|^2 \tag{11}$$

### 3.5. Correlation coefficients representation of time series

When detecting amplitude anomalies is in concern, evaluating the similarity/dissimilarity of time series is performed by using the Euclidean distance function in the original space of time series. On the other hand, for shape anomalies, since the subsequences might be impacted with time shifts, using the Euclidean distance function as a similarity measure in the original feature space is not a suitable choice.

Autocorrelation coefficients of time series have been considered to resolve this issue. By representing subsequences in their autocorrelation coefficients feature space, one may remove time shifts and then the Euclidean distance can be used in the new feature space. As an example, let us consider the generated time series A, B, and C shown in Figure. 4. If we use Euclidean distance for the time series A, B, and C, it is obvious that A and C are more similar (exhibit lower values of the distance) because of the existing time shift between A and B. By considering the autocorrelation space of time series, the time shifts in time series are removed and then time series A and B are considered to be similar.

For a subsequence $\boldsymbol{w}_j$ of length $q$, its autocorrelation coefficients can be calculated in the way (12)

$$a_{j,e} = \frac{\sum_{t=e+1}^{q}(w_{j,t} - \bar{w}_j)(w_{j,t-e} - \bar{w}_j)}{\sum_{t=1}^{q}(w_{j,t} - \bar{w}_j)^2} \tag{12}$$

Where, $e = 1,2,\ldots,q-1$ and $j = 1,2,\ldots,N$. $\bar{w}_j$ is the mean of the time series.

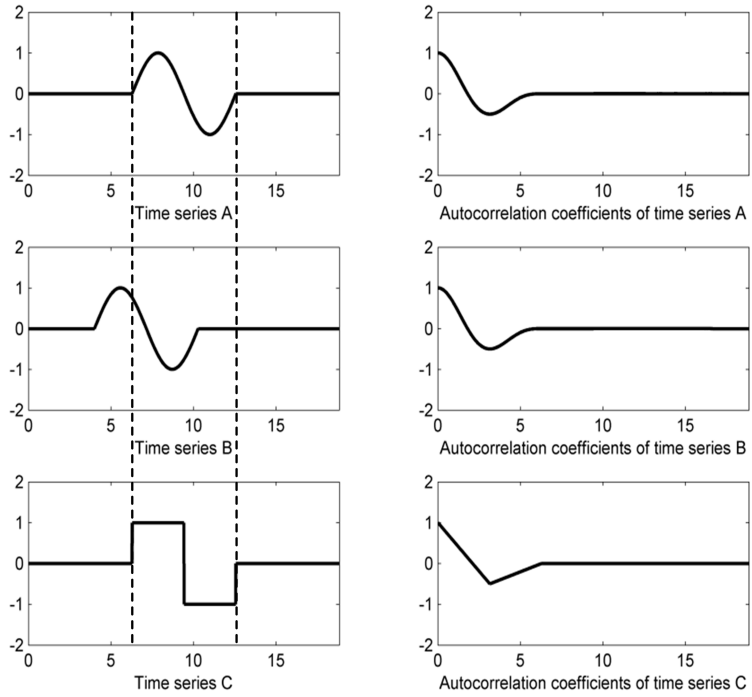

Figure 4 Comparison among subsequences A, B, and C along with their autocorrelation coefficients.

### *3.6. Parameters selection*

Before running the proposed techniques, a Z-score normalization has been applied to each variable (time series) of the multivariate time series to remove the scaling effects among different variables. There are also some parameters such as fuzzy coefficient $m$, number of clusters $c$, the length of subsequences $q$, and the length of each step $r$ whose values can impact the detection process. Comparing the score of anomalous subsequences with the average score of subsequences, a confidence index is considered to optimize those parameters. Figure. 5 shows the idea behind this index when there is only one anomaly in time series.

The confidence index [13] is expressed using the following ratio

$$f = \frac{h_{anomaly}}{\bar{h}} \tag{13}$$

where $h$ stands for the anomaly score for each subsequence, $\bar{h}$ is the average of anomaly scores and $h_{anomaly}$ anomaly score of anomalous subsequences. Encountering a higher value of $f$ indicates assigning a higher anomaly score to anomalous part and lower scores to other parts. As the result, (13) can be used to find optimal values of the parameters. The reason is that higher value of $f$ means that the difference between scores of normal subsequences and abnormal subsequences is bigger. When there is more than one anomaly in time series, $h_{anomaly}$ is the average of anomaly scores of anomalous subsequences.

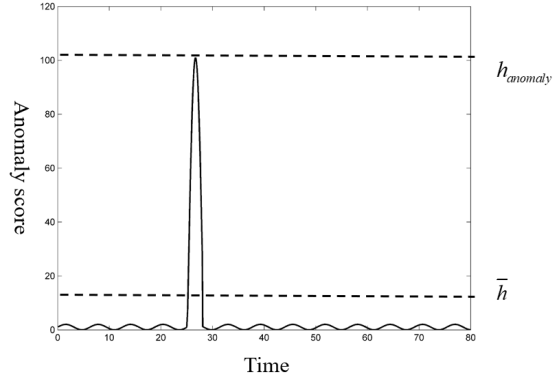

Figure 5 Confidence index (anomaly in the interval [26,29]).

## 4. Experimental Studies

In this section, the clustering-based approach is evaluated in a comprehensive way over a set of synthetic and real-world multivariate time series.

### *4.1. Synthetic datasets*

A synthetic multivariate time series is generated using the ECG simulator with an intent to test the effectiveness of the proposed method [54]. The length of the generated 3-dimensional synthetic time series is 500. The specification is the same as [55]. The individual time series are different because of different heart rates, including 60, 80, and 90. In the series of experiments, the number of clusters was set from 2 to 6. The length of the subsequence and the translation of the sliding window are set to be in the interval [5, 50] and 1, respectively. The optimal of the number of clusters and the length of the subsequence are determined by the confidence index described by (13). The fuzzification coefficient was set to 2.0.

We inserted amplitude anomalies by randomly picking up some windows in different coordinates of the generated multivariate time series and then multiplying the values of picked windows with random values of the interval [0, 3]. Figure. 6 shows the multivariate time series with amplitude anomalies and the anomaly score of each subsequence. It should be noted that this figure only shows the maximal anomaly score for each subsequence because of the overlap between subsequences. When the performance index (13) has been optimized, length of windows and number of clusters are set to 5 and 2, respectively.

The value of the inertia weight w decreases linearly from 0.9 to 0.4. $c_1$ and $c_2$ are 1.49. The maximum number of iterations and the size of the population are 2000 and 500, respectively. The parameter settings of PSO are the same as in the recommendation encountered in [56]. The estimated optimal weights are $\lambda_1 = 0.42$, $\lambda_2 = 0.39$, and $\lambda_3 = 0.19$ when the fitness function is minimized (shown in Fig. 7.). As evidenced in Figure. 6, the proposed method was able to detect all anomalies in the multivariate time series.

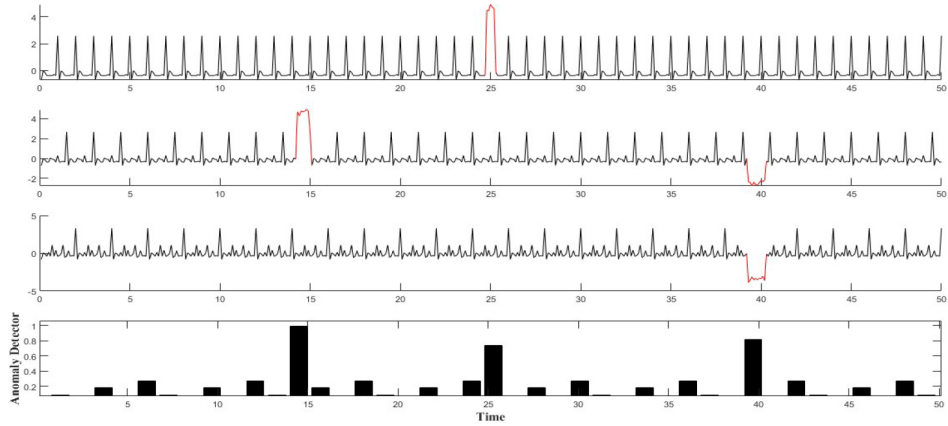

Figure 6 Multivariate time series with existing amplitude anomalies.

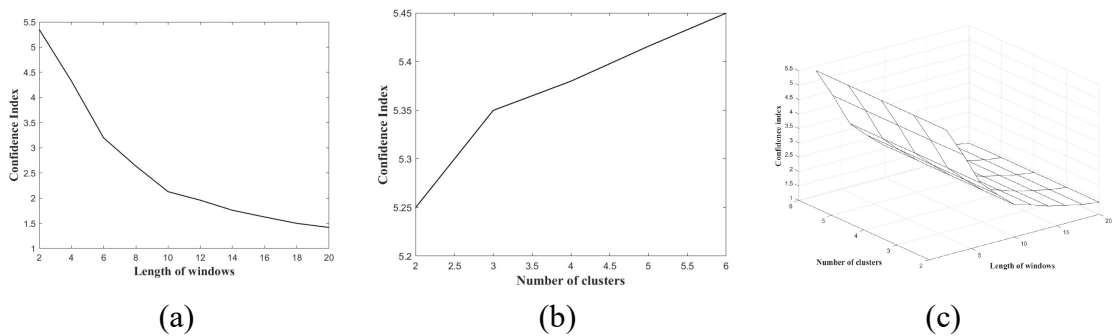

(a)                                        (b)                                        (c)

Figure 7 (a) Different length of windows vs. confidence index (when number of clusters is 2); (b) Different number of clusters vs. confidence index (when length of windows is 5); (c) confidence index when length of sliding window and number of clusters take different values (amplitude anomaly).

Let us now evaluate the proposed method with regard to its abilities to detect and quantify shape anomalies. For this purpose, some shape anomalies have been injected into the generated multivariate time series used in the previous experiment. These shape anomalies are produced by picking up some windows randomly and changing the frequency (by multiplying random value in the interval [1, 3]) of the signal within the windows because frequency change will lead to shape change of the picked windows. Figure. 8 illustrates this multivariate time series with shape anomalies and the obtained corresponding anomaly score for each subsequence determined by the proposed method. The method is able to detect shape anomalies, see Figure. 8.

Figure. 9 illustrates the values of confidence index produced for different number of clusters and the length of subsequences. The best values of these two parameters are 10 and 4, respectively. And the estimated optimal values of the weights are $\lambda_1 = 0.26$, $\lambda_2 = 0.53$, and $\lambda_3 = 0.21$.

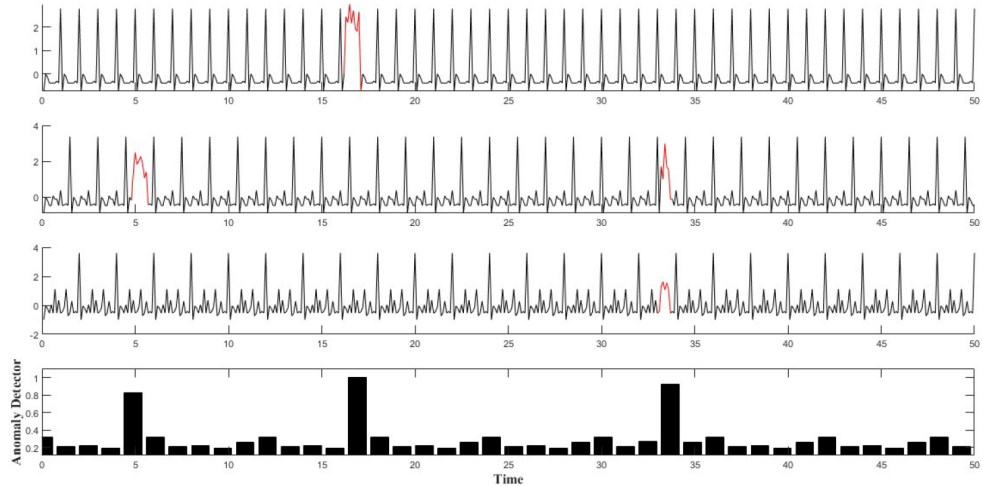

Figure 8 Multivariate time series with existing shape anomalies.

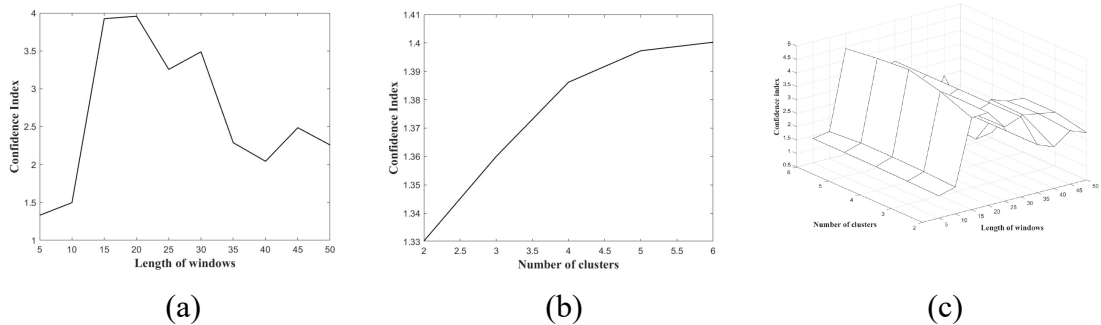

(a)          (b)          (c)

Figure 9 (a) Different length of windows vs. confidence index (when number of clusters is 2); (b) Different number of clusters vs. confidence index (when length of windows is 5); (c) Confidence index when length of sliding window and number of clusters take different values

In the next experiment, a multivariate time series with length 80 is simulated and is shown in Figure. 10. Let us consider we have two time series (two variables) that contain 6 types of different signals (shown in different colors). Each signal is repeated for some times but the number of repeated is unknown while the relationship (1-4, 2-5, 3-6) between different signals are known. Each number represents a specific type of signal. So multivariate time series without anomaly is shown as follows:

**Time series 1**: 1 1 1 1 1 2 2 2 2 2 **2** 3 3 3 3 3

**Time series 2**: 4 4 4 4 4 5 5 5 5 5 **5** 6 6 6 6 6

For the case shown in Figure. 10, multivariate time series with anomaly is shown as follows:

**Time series 1**: 1 1 1 1 1 2 2 2 2 2 **2** 3 3 3 3 3

**Time series 2**: 4 4 4 4 4 5 5 5 5 5 **6** 6 6 6 6 6

Each signal type has been repeated in different positions in each time series. As the result, there is no anomaly in each single time series. However, when considering both time series, there are two anomalies (highlighted) because signal 2 and signal 6 are coming together. Figure. 10 provides the generated time series and the calculated anomaly scores.

As shown in Figure. 10, the proposed method is able to detect the anomalies caused by relations between time series in multivariate time series. The reason is that all univariate subsequences of multivariate subsequence are considered at the same time to determine an anomaly score of this multivariate subsequence.

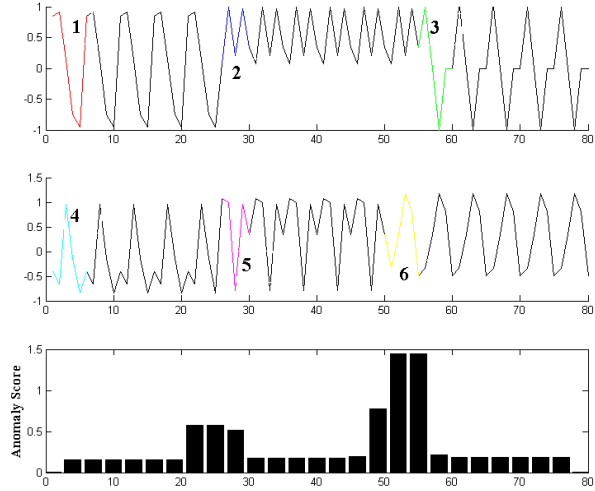

Figure 10 Experimental results of multivariate time series.

In the next experiment, we provide a comparative illustration of the extended FCM and the standard FCM in the proposed general framework of anomaly detection. In order to make all details easily visible, we consider a two-dimensional multivariate time series which consists of four amplitude anomalies (marked by red) as shown in Figure 11 (top). The standard FCM and its extended version were run on this multivariate time series with the same initialization and the following same parameters: number of clusters = 2 and fuzzification coefficient =2.0. Their convergences are archived after a number of iterations. Figure 11 (middle) and (bottom) illustrates the obtained corresponding anomaly scores produced by the anomaly detectors with the extended FCM (with the optimized weights $\lambda_1 = 0.2169$ and $\lambda_1 = 0.7831$) and the standard FCM. The detector with the extended FCM can detect all anomalies while that with the standard FCM cannot work. For an in-depth analysis, we map this multivariate time series onto 2-dimensional space and provide their clustering centers in Figure 12 (a) and (b). They reveal two different data structures within data because an extended version of the Euclidean distance function has been considered to control the impact of each variable in evaluating the similarity between multivariate time series. In addition, there is one error detection (shown in Figure 13 (a) and marked by blue diamond) produced by the detector with the standard FCM. On a basis of different clustering centers, we also provide the corresponding reconstruction version in Figure 13 (b). The distance (or the anomaly score) between actual pattern and its reconstruction version obtained by the extended FCM is less than that obtained by the standard FCM.

As shown in Figure 11-13, the extended FCM is expected to reveal the available data structure for the subsequent anomaly detection because of the different impact of each variable in evaluating the similarity between multivariate time series. The detector with the extended FCM can offer more accurate detection results.

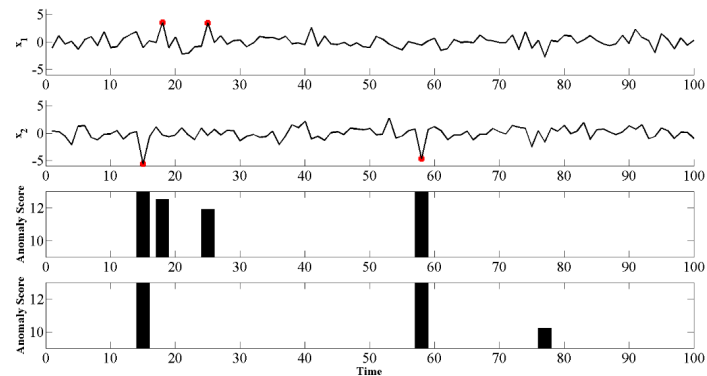

Figure 11 Top: a two-dimensional multivariate time which consists four amplitude anomalies; Middle: experimental results of the extended FCM; Bottom: experimental results of the standard FCM.

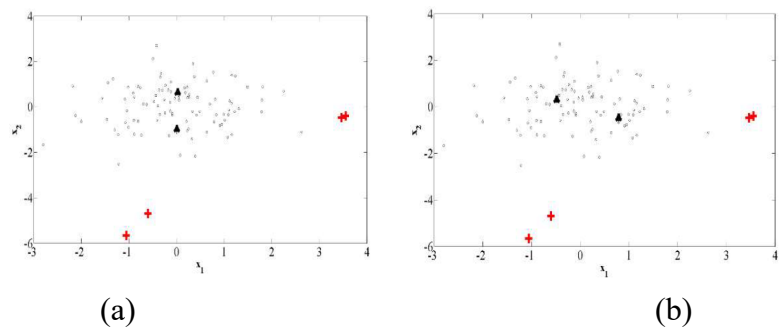

(a)          (b)

Figure 12 (a) Clustering centers (marked by black triangles) obtained by the extended FCM and four amplitude anomalies (marked by red pluses); (b) Clustering centers (marked by black triangles) obtained by the standard FCM and four amplitude anomalies (marked by red pluses);

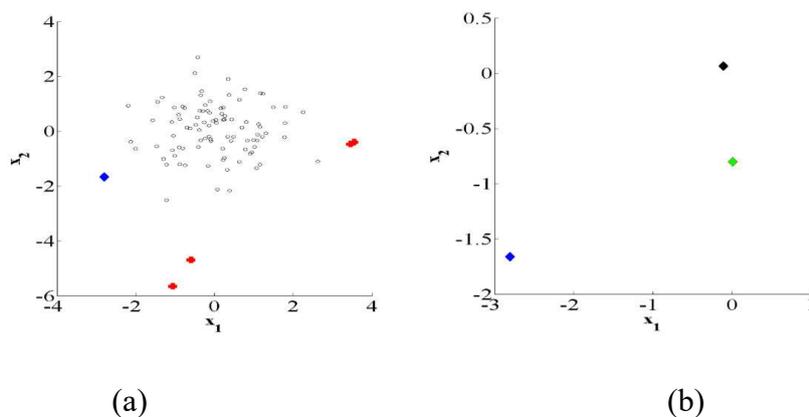

(a)          (b)

Figure 13 (a) anomalies (marked by red pluses) and error detection (marked by blue diamond); (b) error detection (marked by blue diamond) and its reconstruction versions based on the extended FCM and the standard FCM respectively.

### *4.2. Publicly available datasets*

In this subsection, two real-world datasets, namely electrocardiogram dataset, and climate change dataset have been studied for anomaly detection in shape and amplitude using the proposed method.

*4.2.1 Electrocardiograms (ECG) dataset*

This dataset is obtained from the MIT-BIH Arrhythmia database [56]. In the ECG dataset, each heartbeat is annotated by two cardiologists independently so it can be used as a benchmark to evaluate the proposed method.

In this experiment, three ECG heartbeats with some annotated anomalies are considered. To consider all type of beats in a time window, the length of time window has been considered to be 1.2 times average length of the RR interval [13]. The other parameters are selected in a similar fashion as used in the previous experiments. The optimal values of the number of clusters and the length of subsequence are determined by the confidence index mentioned in (13). The fuzzification coefficient was set to 2.0 and the length of each slide (move) of the sliding window is set to 10% of the length of subsequences. Figure. 14 illustrates the ECG signals and the corresponding anomaly scores generated by the proposed method. The anomalous part of each dataset is highlighted. The detected anomalous parts of data correspond to premature ventricular contraction (marked in red color) or Atrial premature contraction (marked in green). In all cases, the method can find the abnormal parts of multivariate time series. In Figure. 14 (b), one of the detected anomalies is annotated as normal by cardiologists (highlighted in magenta), however this part of data has visible difference in shape when compared with other normal portions of the series. As can be seen in Figure. 14, shape information of anomalous parts (highlighted in different color) is different from that of normal parts of multivariate time series. The difference will lead to different clustering results and yields a higher anomaly score of multivariate subsequences.

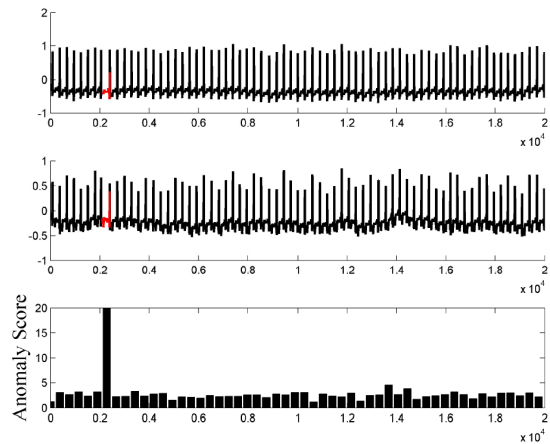

(a)

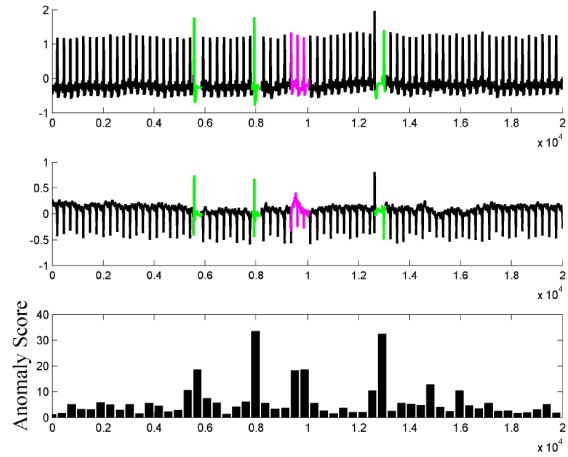

(b)

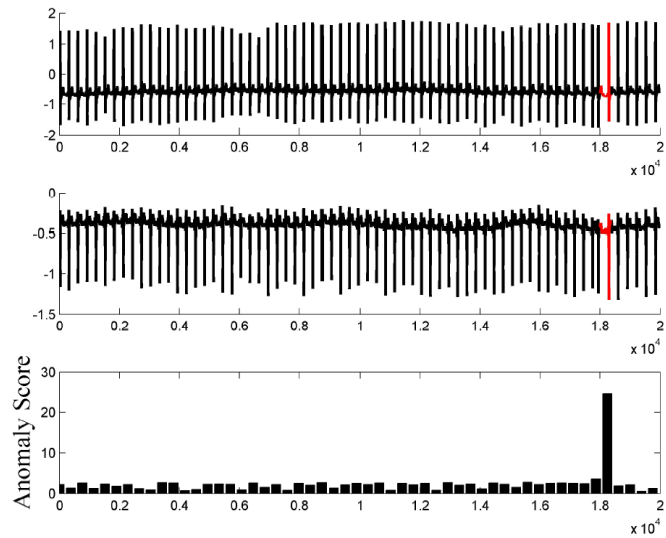

(c)

Figure 14 MIT-BIH arrhythmia datasets.

*4.2.2 Climate change dataset*

In this section, we consider the monthly measurements of climatological factors from January 1990 to December 2002, [47] in three different places of USA. They contain vapor (VAP), temperature (TMP), temperature min (TMN), temperature max (TMX), global solar radiation (GLO), extraterrestrial radiation (ETR) and extraterrestrial normal radiation (ETRN) from CRU (http://www.cru.uea.ac.uk/cru/data), NOAA (http://www.esrl.noaa.gov/gmd/dv/ftpdata.html), and NCDC (http://rredc.nrel.gov/solar/old_data/nsrdb/).

Some simulated amplitude and shape anomalies are inserted into this multivariate time series to test our anomaly detection technique. For instance, the values of global solar radiation in the first place from April 1995 to November 1996 are doubled. The values of temperature recorded at the second place from January 1997 to December 1997 are replaced by that of July 1990 to June 1991. A sliding window of length 12 is considered to reflect successive months. The other parameter settings are similar to those used in the previous experiments. The fuzzification coefficient was set to 2.0 and the number of clusters was varied from 2 to 6. The length of each movement of the sliding window is equal to 10% of the length of the subsequences. The optimal values of weights are shown in Table 1. Figure. 15 illustrates the experimental results showing the detected anomalous parts.

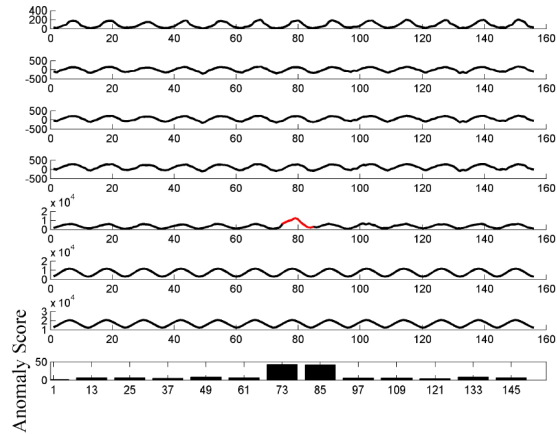

(a)

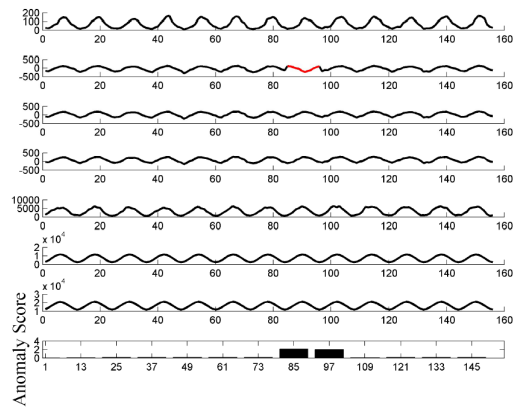

(b)

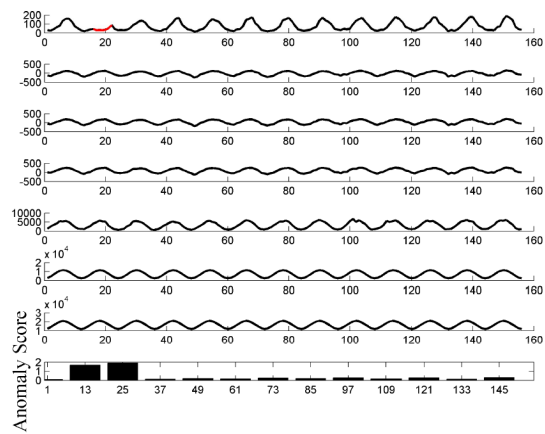

(c)

Figure 15 MIT-BIH arrhythmia dataset.

Table 1  Optimal values of weights

| climate change data | $\lambda_1$ | $\lambda_2$ | $\lambda_3$ | $\lambda_4$ | $\lambda_5$ | $\lambda_6$ | $\lambda_7$ |
|---|---|---|---|---|---|---|---|
| (a) | 0.02 | 0.20 | 0.26 | 0.04 | 0.23 | 0.03 | 0.22 |
| (b) | 0.12 | 0.16 | 0.08 | 0.15 | 0.16 | 0.16 | 0.17 |
| (c) | 0.14 | 0.08 | 0.01 | 0.16 | 0.00 | 0.32 | 0.29 |

From the above table, we can infer that the impact of different variables could be different in the clustering process even if similar datasets are used. The higher the weights are, the more important the variables are in the clustering process.

*4.2.3 A comparison with distance-based methods*

In a general framework of distance-based anomaly detection techniques one calculates, the similarity of each subsequence with all other subsequences in the dataset and selects the subsequences with lowest similarity (highest distance) to other subsequences as anomalous parts of data. A 1-NN technique has been used for this purpose. The main drawback of 1-NN based anomaly detection methods is that if two similar anomalies happen in a time series, the technique is not able to detect them. To resolve this problem, one may consider the use of the *K*-NN classifier. However, finding the parameter *K* is a challenging problem. On the other hand, in clustering-based methods, by considering the number of clusters equal to 2 or 3 (or using a cluster validity index approach) one may detect anomalies even if some anomalies are similar.

Here we compare the proposed clustering-based technique with the one discussed in [1]. A two-dimensional multivariate time series is constructed and shown in Figure. 16. Two anomalies are inserted into this multivariate time series. It is noticeable that the distance-based method is not able to detect all anomalous parts of data.

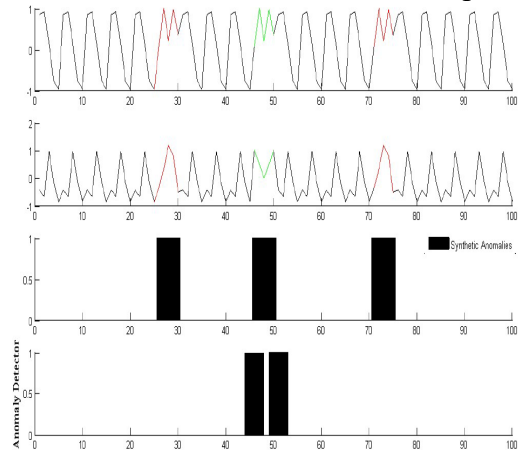

(a)

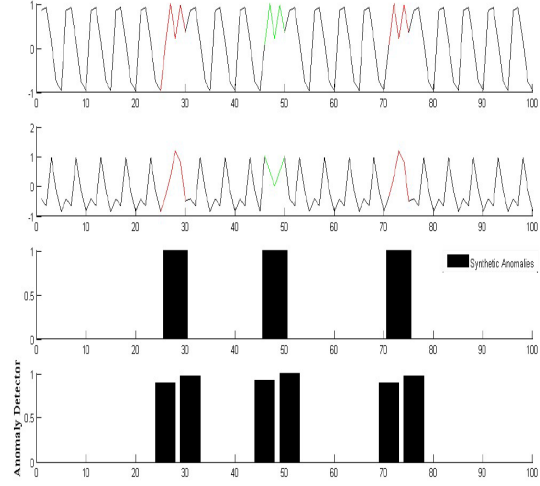
(b)

Figure 16 The proposed method vs. a 1-NN technique: (a) experimental result of 1-NN method; (b) experimental result of the proposed method.

*4.2.4 A comparison with model-based methods*

In this subsection, we report the comparison results between the proposed method and the existing methods on various real-world multivariate time series from the UCI machine learning repository [57] and DataMarket [60]. Three real-world datasets are used: U.S. Dollar Exchange Rate, EEG Eye State, and Air Quality Dataset. U.S. Dollar Exchange Rate dataset contains historical intraday data (i.e., Dutch guilder, French franc, and German mark) from January 03, 1989, to December 31, 1988. EEG Eye State Dataset contains electroencephalogram measurements (i.e., AF3, F7, and FC5). Air Quality Dataset contains different hourly average concentrations (i.e., Total Nitrogen Oxide, Nitrogen Dioxide, and Ozone) generated by the air quality sensors in an Italian city from March 10, 2004, to June 2 2004. We used the same parameter setting as presented before. For a clearer comparison, quantitative evaluation is also included. TP, FP, TN, and FN are True Positive, False Positive, True Negative, and False Negative, respectively. The Accuracy, F-measure, Sensitivity, and Specificity are given as follows.

$$\text{Accuracy} = \frac{\text{TN}+\text{TP}}{\text{TN}+\text{FP}+\text{FN}+\text{TP}}$$

$$\text{Sensitivity} = \frac{\text{TP}}{\text{TP}+\text{FN}}$$

$$\text{Specificity} = \frac{\text{TN}}{\text{TN}+\text{FP}}$$

$$\text{F-measure} = \frac{2 \times \text{Precision} \times \text{Recall}}{\text{Precision} \times \text{Recall}}$$

Where

$$\text{Precision} = \frac{\text{TP}}{\text{TP}+\text{FP}} \quad and \quad \text{Recall} = \frac{\text{TP}}{\text{TP}+\text{FN}}$$

Since the outputs of our approaches are anomaly scores instead of the statuses, we need to convert all anomaly scores to 'normal' or 'abnormal' status using thresholds. Figure 17–19 describe the comparison results of PCA (Principal Component

Analysis)+HMM(Hidden Markov Model), FCM+HMM, Sugeno Integral+HMM, Choquet Integral+HMM, and the proposed method when the threshold takes different values.

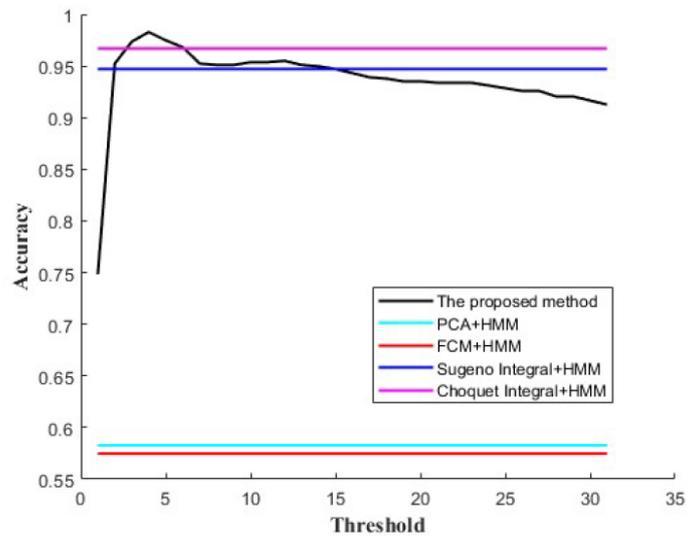

(a)

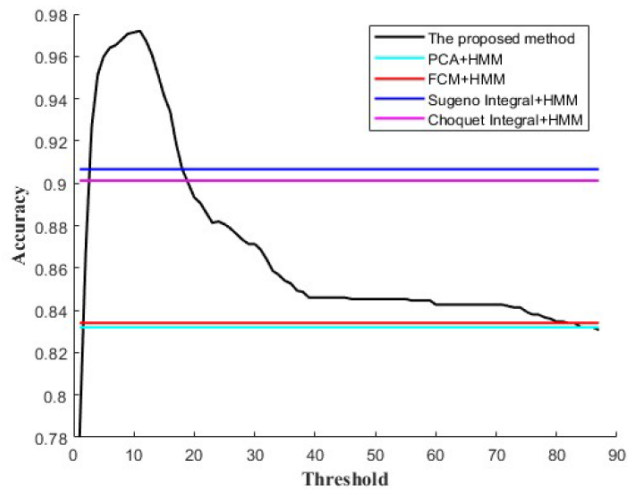

(b)

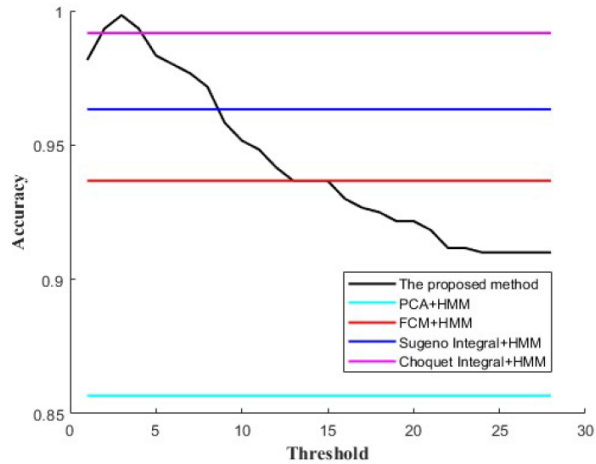

(c)

Figure 17 Accuracy for different threshold values: (a) US Dollar exchange rate, (b) EEG eye state, (c)Air quality

Selecting the suitable threshold, which is our future work, is important for the conversation between anomaly scores and the status (e.g., 'Normal' and 'Abnormal'), as shown in Figure 17. In this study, we used the threshold value when the detection accuracy is maximum. Figure 18-20 illustrate the comparison among the existing methods and the proposed methods on testing sets of different datasets.

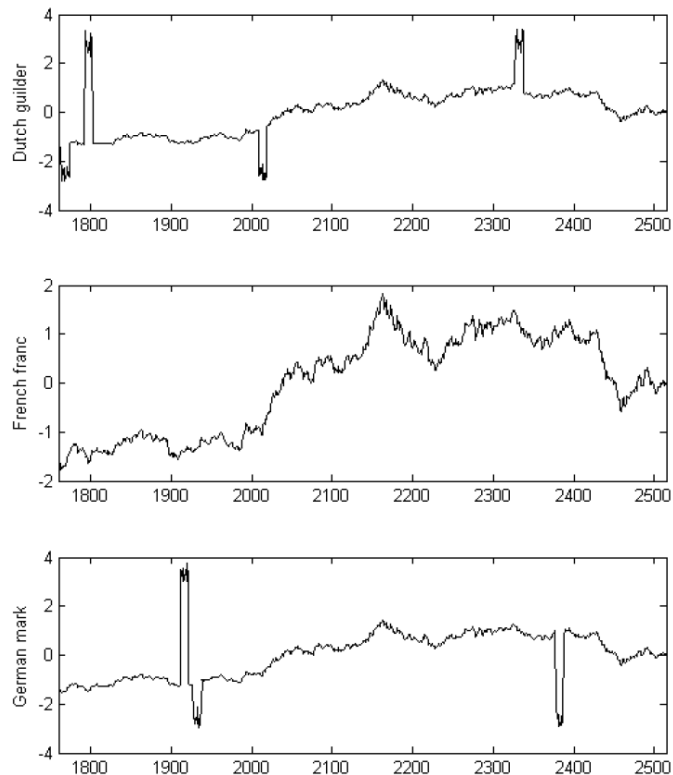

(a)

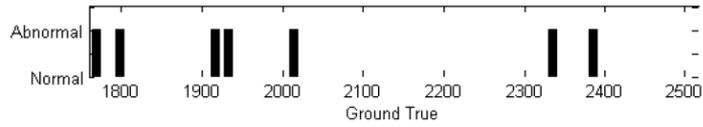
(b)

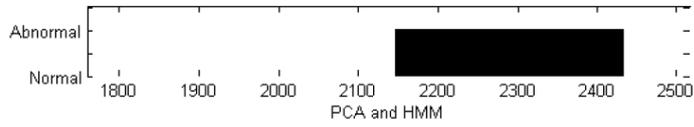
(c)

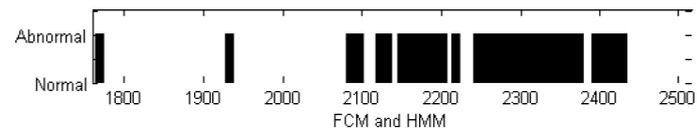
(d)

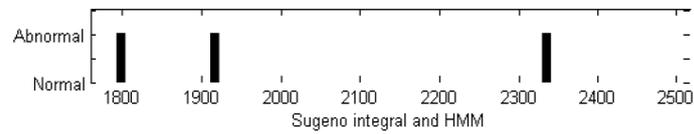
(e)

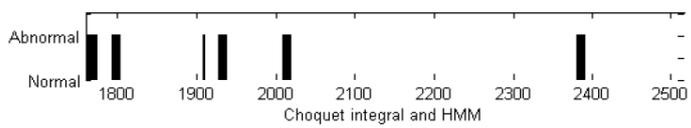
(f)

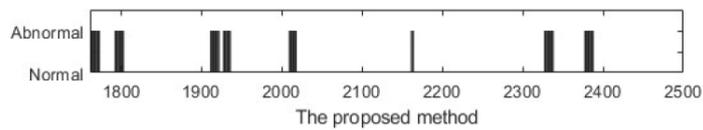
(g)

Figure 18 U S. Dollar Exchange Rate Dataset: (a) testing set, (b) Ground truth (c) Experimental results of PCA+ HMM, (d) Experimental results of FCM + HMM, (e) Experimental results of Sugeno integral + HMM, (f) Experimental results of Choquet integral + HMM, (g) Experimental results of the proposed method.

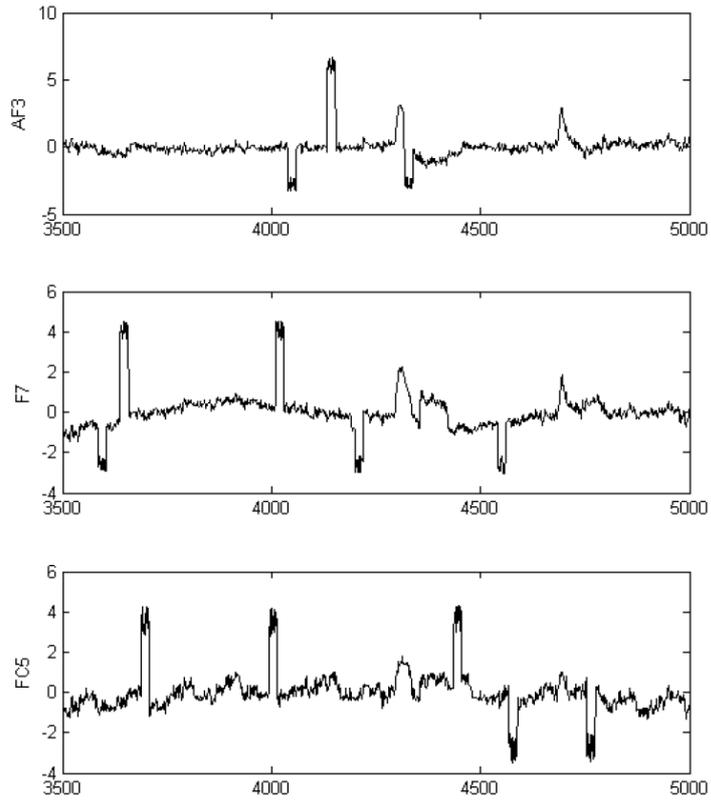

(a)

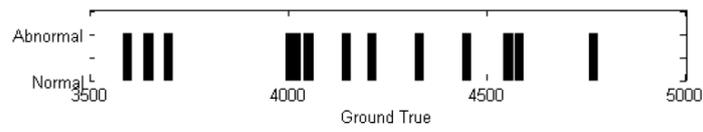

(b)

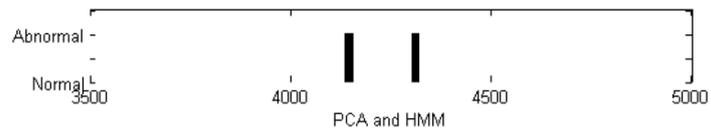

(c)

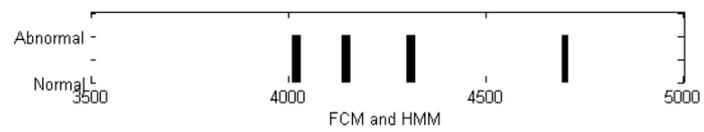

(d)

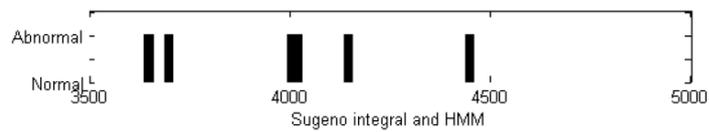

(e)

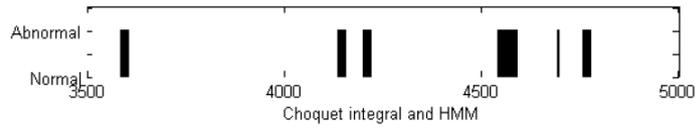

(f)

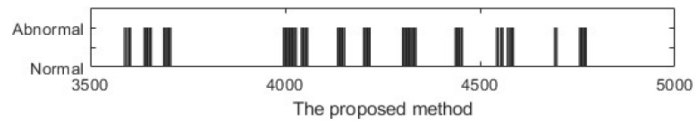

(g)

Figure 19 EEG Eye State Dataset: (a) testing set, (b) Ground truth (c) Experimental results of PCA+ HMM, (d) Experimental results of FCM + HMM, (e) Experimental results of Sugeno integral + HMM, (f) Experimental results of Choquet integral + HMM, (g) Experimental results of the proposed method.

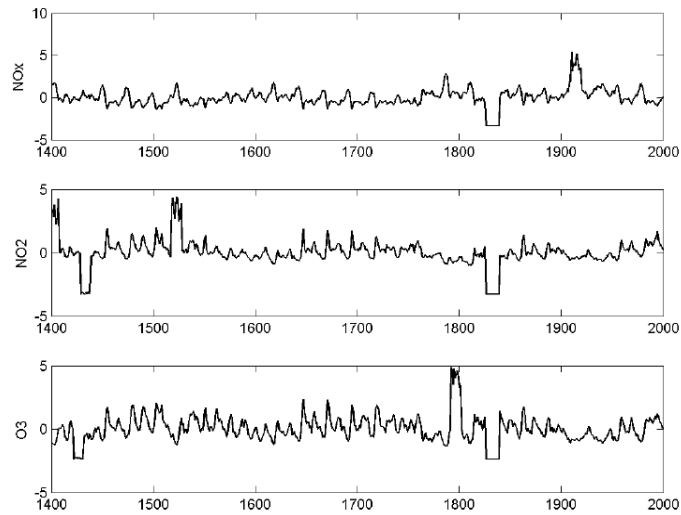

(a)

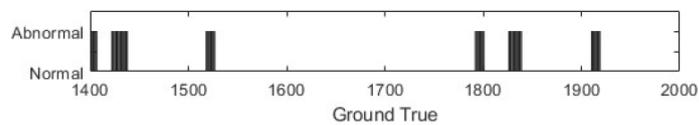

(b)

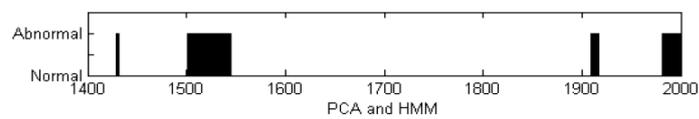

(c)

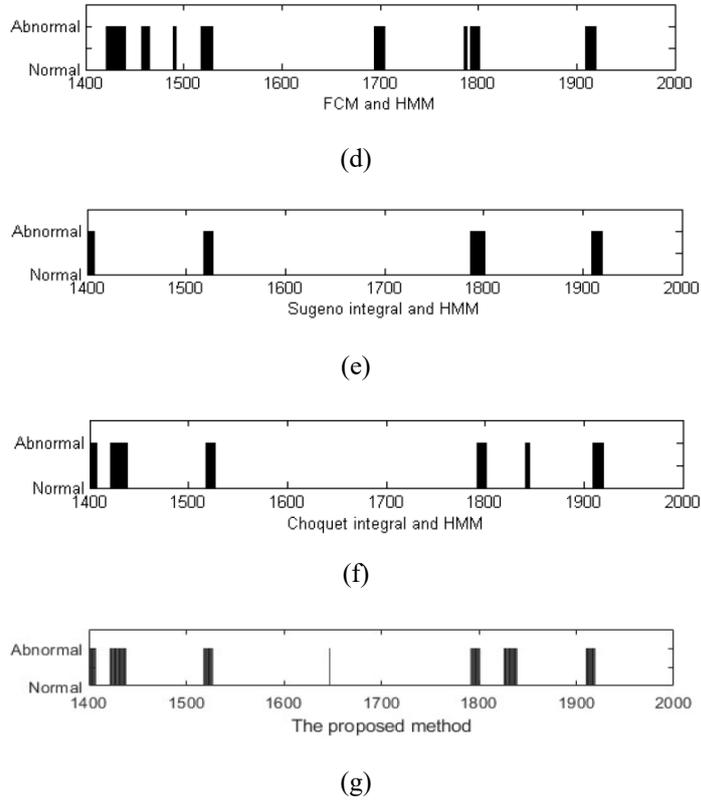

(d)

(e)

(f)

(g)

Figure 20 Air Quality Dataset: (a) testing set, (b) Ground truth (c) Experimental results of PCA+ HMM, (d) Experimental results of FCM + HMM, (e) Experimental results of Sugeno integral + HMM, (f) Experimental results of Choquet integral + HMM, (g) Experimental results of the proposed method.

Table 2-4 summarize their objective performance comparison by using different evaluation criterions such as accuracy, sensitivity, specificity, and F-measure.

Table 2 Experimental results in U.S. Dollar Exchange Rate dataset

| Methods | Accuracy | Sensitivity | Specificity | F-measure |
| --- | --- | --- | --- | --- |
| PCA+HMM | 0.5828 | 0.6131 | 0.2857 | 0.7272 |
| FCM+HMM | 0.5748 | 0.5898 | 0.4286 | 0.7156 |
| Sugeno Integral+HMM | 0.947 | 1 | 0.4286 | 0.9716 |
| Choquet Integral+HMM | 0.9669 | 0.9927 | 0.7143 | 0.982 |
| The proposed method | 0.9828 | 1 | 0.8434 | 0.9904 |

Table 3 Experimental results in EEG Eye State dataset

| Methods | Accuracy | Sensitivity | Specificity | F-measure |
| --- | --- | --- | --- | --- |
| PCA+HMM | 0.832 | 0.9879 | 0.0742 | 0.907 |
| FCM+HMM | 0.834 | 0.9735 | 0.1563 | 0.9068 |
| Sugeno Integral+HMM | 0.9067 | 1 | 0.4531 | 0.9468 |
| Choquet Integral+HMM | 0.9013 | 0.9904 | 0.4688 | 0.9434 |
| The proposed method | 0.972 | 0.9959 | 0.8715 | 0.9829 |

Table 4 Experimental results in Air Quality Dataset

| Methods | Accuracy | Sensitivity | Specificity | F-measure |
|---|---|---|---|---|
| PCA+HMM | 0.8567 | 0.9048 | 0.3704 | 0.92 |
| FCM+HMM | 0.9367 | 0.9469 | 0.8333 | 0.9646 |
| Sugeno Integral+HMM | 0.9633 | 0.9908 | 0.6852 | 0.98 |
| Choquet Integral+HMM | 0.9917 | 0.9908 | 1 | 0.9954 |
| The proposed method | 0.9983 | 1 | 0.9855 | 0.9991 |

As illustrated in Figure 19-22 and Table 2-4, the performance of the proposed methods is better because the extended fuzzy clustering technique can reveal the data structure by using the optimal weights, shown in Table 5.

Table 5 The optimal values of $\lambda$ in different datasets

| Dataset | $\lambda_1$ | $\lambda_2$ | $\lambda_3$ |
|---|---|---|---|
| U.S. Dollar Exchange Rate dataset | 0.68 | 0.09 | 0.23 |
| EEG Eye State dataset | 0.24 | 0.17 | 0.59 |
| Air Quality dataset | 0.41 | 0.31 | 0.28 |

## 5. Conclusions

In this study, we have introduced novel methods of amplitude and shape anomaly detection for multivariate time series. The extended version of Fuzzy C-Means is used to capture the structure of multivariate time series by achieving a balance among the effect of different components in the clustering process. The reconstruction of the multivariate subsequence is performed by using the revealed structure. A reconstruction error serves as the fitness function of the PSO algorithm and has been considered as the level of anomaly detected in each subsequence. We conducted experiments using both synthetic and real-world datasets and compared the proposed methods with several techniques reported in the literature. Experimental results demonstrate the effectiveness of the proposed methods to detect amplitude and shape anomalies in multivariate time series.

From a practical point of view, the proposed clustering-based anomaly detection methods help the user exploit the large amounts of unlabeled high dimensional time series that are available. Different from those supervised multivariate time series anomaly detection methods, the proposed methods do not need the label (i.e., normal, or abnormal) of each timestamp for the models training. The labels of multivariate time series are often generated manually by exports, which is often low efficiency, time-consuming, and high cost. However, detecting anomalies in a limited time is a matter of particular importance for real-world applications, which generate vast volumes of data in an automated way and datasets with missing or incorrect labels.

Although the proposed method shows good performance, it comes with some limitations. One limitation of the proposed methods is that it is not guaranteed that the structure of multivariate time series can be revealed when the dimension of multivariate time series is high because of the possibility that the PSO falls into local optima and the drawback of FCM. Another limitation is an intensive computing overhead resulting from the fact that the weighted Euclidean distance function calls for a substantial level of computing because of the time-consuming weights learning process. Considering the computing overhead associated with the analysis of different combinations of the entries of one may consider a simplified version of the method with some predetermined values of the entries of the vector of these parameters.

In the future, we plan to add the Fourier transform or wavelet transform based feature extraction models to improve the performance of the proposed framework because they can convert the time series from the time domain to frequency domain. Future work will also consider the introduction of kernel methods to reveal the structure of the data with a more complex structure. Next, another future direction is to compare the performance of different evolutionary algorithms (e.g., Genetic Algorithms GA, Flock of Starlings Optimization FSO and their variants) in the proposed framework. The essential idea of Genetic Algorithms is to maintain a population of potential solutions which are evaluated, selected, crossed, mutated and replaced to perform optimization. Being different from PSO that is suitable for carrying out search in the continuous space, GA comes with evolution operators such as crossover/mutation and usually works more efficiently with discrete search spaces. In addition, one has to take into consideration computational overhead before selecting the optimizer for real-world applications. Generally, the larger the number of variables and fitness function evaluations are, the more computational resources are required to achieve the optimization goal. With regard to computing resources, a statistical comparison of different optimizers has been provided in [58]. Since the PSO parameters impacts the performance of the method, future studies could focus on selecting their suitable values as well as considering settings-free versions of the PSO. Although the classical PSO along with the use of its static parameters delivers satisfactory results because of its simplicity and relatively high convergence speed, the variants of PSO without the need for any detailed refinements of the setting could be considered as a further extension of this work. This direction could be attractive given that there are more complicated datasets arising in the real-world applications (e.g., finance, geographic, biology and telecommunication). When handling the high-dimensional and large-scale real-world datasets characterized by additional complex features, the setting-free PSO versions are expected to perform better because of their self-tuning variants. Moreover, selecting a suitable threshold value of the anomaly score to discriminate between normal and abnormal segments of time series is of evident interest.

## Acknowledgments
This work was supported by the National Science Centre, Poland, within the framework of the project no. 2017/25/B/ST6/00114.